\documentclass[runningheads]{llncs}

\usepackage{eccv}

\usepackage{eccvabbrv}

\usepackage{graphicx}
\usepackage{booktabs}
\usepackage{multirow}
\usepackage{multicol}
\usepackage{kotex}

\usepackage[accsupp]{axessibility}  

\usepackage{tikz,subcaption}
\usetikzlibrary{spy,backgrounds} 
\usepackage{xcolor}
\usepackage{amsmath}
\usepackage{algorithm}
\usepackage{algpseudocode}
\usepackage{nicefrac}
\usepackage{indentfirst}
\usepackage{tabularx}
\usepackage{color}
\usepackage{colortbl}
\usepackage{graphicx}
\usepackage{lipsum}
\usepackage{wrapfig}
\usepackage{graphicx}
\usepackage{caption}
\usepackage{subcaption}
\usepackage{tikz}
\usepackage{pgfplots}
\usetikzlibrary{spy,calc}
\usepackage{hyperref}
\newif\ifblackandwhitecycle
\gdef\patternnumber{0}
\pgfkeys{/tikz/.cd,
    zoombox paths/.style={
        draw=orange,
        very thick
    },
    black and white/.is choice,
    black and white/.default=static,
    black and white/static/.style={ 
        draw=white,   
        zoombox paths/.append style={
            draw=white,
            postaction={
                draw=black,
                loosely dashed
            }
        }
    },
    black and white/static/.code={
        \gdef\patternnumber{1}
    },
    black and white/cycle/.code={
        \blackandwhitecycletrue
        \gdef\patternnumber{1}
    },
    black and white pattern/.is choice,
    black and white pattern/0/.style={},
    black and white pattern/1/.style={    
            draw=white,
            postaction={
                draw=black,
                dash pattern=on 2pt off 2pt
            }
    },
    black and white pattern/2/.style={    
            draw=white,
            postaction={
                draw=black,
                dash pattern=on 4pt off 4pt
            }
    },
    black and white pattern/3/.style={    
            draw=white,
            postaction={
                draw=black,
                dash pattern=on 4pt off 4pt on 1pt off 4pt
            }
    },
    black and white pattern/4/.style={    
            draw=white,
            postaction={
                draw=black,
                dash pattern=on 4pt off 2pt on 2 pt off 2pt on 2 pt off 2pt
            }
    },
    zoomboxarray inner gap/.initial=5pt,
    zoomboxarray columns/.initial=2,
    zoomboxarray rows/.initial=2,
    subfigurename/.initial={},
    figurename/.initial={zoombox},
    zoomboxarray/.style={
        execute at begin picture={
            \begin{scope}[
                spy using outlines={%
                    zoombox paths,
                    width=\imagewidth / \pgfkeysvalueof{/tikz/zoomboxarray columns} - (\pgfkeysvalueof{/tikz/zoomboxarray columns} - 1) / \pgfkeysvalueof{/tikz/zoomboxarray columns} * \pgfkeysvalueof{/tikz/zoomboxarray inner gap} -\pgflinewidth,
                    height=\imageheight / \pgfkeysvalueof{/tikz/zoomboxarray rows} - (\pgfkeysvalueof{/tikz/zoomboxarray rows} - 1) / \pgfkeysvalueof{/tikz/zoomboxarray rows} * \pgfkeysvalueof{/tikz/zoomboxarray inner gap}-\pgflinewidth,
                    magnification=3,
                    every spy on node/.style={
                        zoombox paths
                    },
                    every spy in node/.style={
                        zoombox paths
                    }
                }
            ]
        },
        execute at end picture={
            \end{scope}
            \node at (image.north) [anchor=north,inner sep=0pt] {\subcaptionbox{\label{\pgfkeysvalueof{/tikz/figurename}-image}}{\phantomimage}};
            \node at (zoomboxes container.north) [anchor=north,inner sep=0pt] {\subcaptionbox{\label{\pgfkeysvalueof{/tikz/figurename}-zoom}}{\phantomimage}};
     \gdef\patternnumber{0}
        },
        spymargin/.initial=0.5em,
        zoomboxes xshift/.initial=1,
        zoomboxes right/.code=\pgfkeys{/tikz/zoomboxes xshift=1},
        zoomboxes left/.code=\pgfkeys{/tikz/zoomboxes xshift=-1},
        zoomboxes yshift/.initial=0,
        zoomboxes above/.code={
            \pgfkeys{/tikz/zoomboxes yshift=1},
            \pgfkeys{/tikz/zoomboxes xshift=0}
        },
        zoomboxes below/.code={
            \pgfkeys{/tikz/zoomboxes yshift=-1},
            \pgfkeys{/tikz/zoomboxes xshift=0}
        },
        caption margin/.initial=4ex,
    },
    adjust caption spacing/.code={},
    image container/.style={
        inner sep=0pt,
        at=(image.north),
        anchor=north,
        adjust caption spacing
    },
    zoomboxes container/.style={
        inner sep=0pt,
        at=(image.north),
        anchor=north,
        name=zoomboxes container,
        xshift=\pgfkeysvalueof{/tikz/zoomboxes xshift}*(\imagewidth+\pgfkeysvalueof{/tikz/spymargin}),
        yshift=\pgfkeysvalueof{/tikz/zoomboxes yshift}*(\imageheight+\pgfkeysvalueof{/tikz/spymargin}+\pgfkeysvalueof{/tikz/caption margin}),
        adjust caption spacing,
    },
    calculate dimensions/.code={
        \pgfpointdiff{\pgfpointanchor{image}{south west} }{\pgfpointanchor{image}{north east} }
        \pgfgetlastxy{\imagewidth}{\imageheight}
        \global\let\imagewidth=\imagewidth
        \global\let\imageheight=\imageheight
        \gdef\columncount{1}
        \gdef\rowcount{1}
        
    },
    image node/.style={
        inner sep=0pt,
        name=image,
        anchor=south west,
        append after command={
            [calculate dimensions]
            node [image container,subfigurename=\pgfkeysvalueof{/tikz/figurename}-image] {\phantomimage}
            node [zoomboxes container,subfigurename=\pgfkeysvalueof{/tikz/figurename}-zoom] {\phantomimage}
        }
    },
    color code/.style={
        zoombox paths/.append style={draw=#1}
    },
    connect zoomboxes/.style={
    spy connection path={\draw[draw=none,zoombox paths] (tikzspyonnode) -- (tikzspyinnode);}
    },
    help grid code/.code={
        \begin{scope}[
                x={(image.south east)},
                y={(image.north west)},
                font=\footnotesize,
                help lines,
                overlay
            ]
            \foreach \x in {0,1,...,9} { 
                \draw(\x/10,0) -- (\x/10,1);
                \node [anchor=north] at (\x/10,0) {0.\x};
            }
            \foreach \y in {0,1,...,9} {
                \draw(0,\y/10) -- (1,\y/10);                        \node [anchor=east] at (0,\y/10) {0.\y};
            }
        \end{scope}    
    },
    help grid/.style={
        append after command={
            [help grid code]
        }
    },
}
\newcommand\phantomimage{%
    \phantom{%
        \rule{\imagewidth}{\imageheight}%
    }%
}
\newcommand\zoombox[2][]{
    \begin{scope}[zoombox paths]
        \pgfmathsetmacro\xpos{
            (\columncount-1)*(\imagewidth / \pgfkeysvalueof{/tikz/zoomboxarray columns} + \pgfkeysvalueof{/tikz/zoomboxarray inner gap} / \pgfkeysvalueof{/tikz/zoomboxarray columns} ) + \pgflinewidth
        }
        \pgfmathsetmacro\ypos{
            (\rowcount-1)*( \imageheight / \pgfkeysvalueof{/tikz/zoomboxarray rows} + \pgfkeysvalueof{/tikz/zoomboxarray inner gap} / \pgfkeysvalueof{/tikz/zoomboxarray rows} ) + 0.5*\pgflinewidth
        }
        \edef\dospy{\noexpand\spy [
            #1,
            zoombox paths/.append style={
                black and white pattern=\patternnumber
            },
            every spy on node/.append style={#1},
            x=\imagewidth,
            y=\imageheight
        ] on (#2) in node [anchor=north west] at ($(zoomboxes container.north west)+(\xpos pt,-\ypos pt)$);}
        \dospy
        \pgfmathtruncatemacro\pgfmathresult{ifthenelse(\columncount==\pgfkeysvalueof{/tikz/zoomboxarray columns},\rowcount+1,\rowcount)}
        \global\let\rowcount=\pgfmathresult
        \pgfmathtruncatemacro\pgfmathresult{ifthenelse(\columncount==\pgfkeysvalueof{/tikz/zoomboxarray columns},1,\columncount+1)}
        \global\let\columncount=\pgfmathresult
        \ifblackandwhitecycle
            \pgfmathtruncatemacro{\newpatternnumber}{\patternnumber+1}
            \global\edef\patternnumber{\newpatternnumber}
        \fi
    \end{scope}
}

\usepackage{multirow}
\usepackage[final]{changes}     %
\usepackage{booktabs}
\usepackage{makecell}
\usepackage{float}

\newcommand{\titleshort}{ObsGraph}

\usepackage{hyperref}

\usepackage{orcidlink}

\begin{document}

\title{ObsGraph: Hierarchical Observation Representation for Embodied Reasoning and Exploration} 

\titlerunning{ObsGraph}

\author{Taekbeom Lee,
Youngseok Jang, \\
Jeonghwa Heo,
Jeongjun Choi,
H. Jin Kim}

\authorrunning{T.~Lee et al.}

\institute{
Seoul National University\\
\email{ltb1128@snu.ac.kr}
}

\maketitle

\begin{abstract}
Embodied reasoning and exploration are increasingly considered crucial abilities for robots operating in complex and unfamiliar environments.
To accomplish tasks in such settings, an agent must identify and acquire the information necessary for the task through exploration.
We propose \titleshort{}, an observation-centric hierarchical scene graph that unifies scene representation, retrieval, and exploration.
It retains visual evidence and organizes it into room–view–object layers: rooms provide coarse semantic anchors, views preserve contextual object co-visibility, and objects store fine-grained details. On top of this representation, we perform coarse-to-fine hierarchical retrieval under a bounded budget, and crucially use retrieval outcomes to structure the exploration candidate space—activating room-level exploration, view refinement, or frontier exploration—thereby tightly coupling representation, retrieval, and adaptive multi-scale exploration. Experiments across embodied reasoning and exploration benchmarks demonstrate improved success and efficiency, highlighting the benefits of structured scene representation and more targeted information gathering driven by identified evidence gaps.
\end{abstract}

\section{Introduction}
\label{sec:intro}

Autonomous robots are expected to operate as versatile assistants in human environments, with stronger capabilities for perception \cite{zhang2023clip, li2022blip, antol2015vqa} and interaction \cite{ahn2022can, brohan2022rt, zitkovich2023rt}.
Both of these capabilities often assume the targets are fully observable.
However, in the real world, robots are often not aware of the environment, and they should move and search to find relevant targets to meet task specifications.
To complete such tasks efficiently and reliably, robots must reason about what is known, identify what is missing, and actively gather the necessary information.

In this context, embodied reasoning and exploration are crucial, especially in complex environments.
This capability refers to a robot's ability to actively explore a scene, reason over accumulated observations, and identify missing information required for task completion.
Its effectiveness fundamentally depends on scene representation, namely, how the agent organizes and maintains observations over time.
As tasks vary, including multiple objectives within the same environment, the representation should be structured in a task-agnostic manner while preserving diverse evidence and supporting flexible access to task-relevant information.
At the same time, the agent's understanding of the environment is inherently partial, given that observations are limited by the exploration history. 
Therefore, the representation should make explicit what is currently known with respect to the given task and where additional observation is needed.

\begin{figure}[t!]
    \centering
    \includegraphics[width=\textwidth]{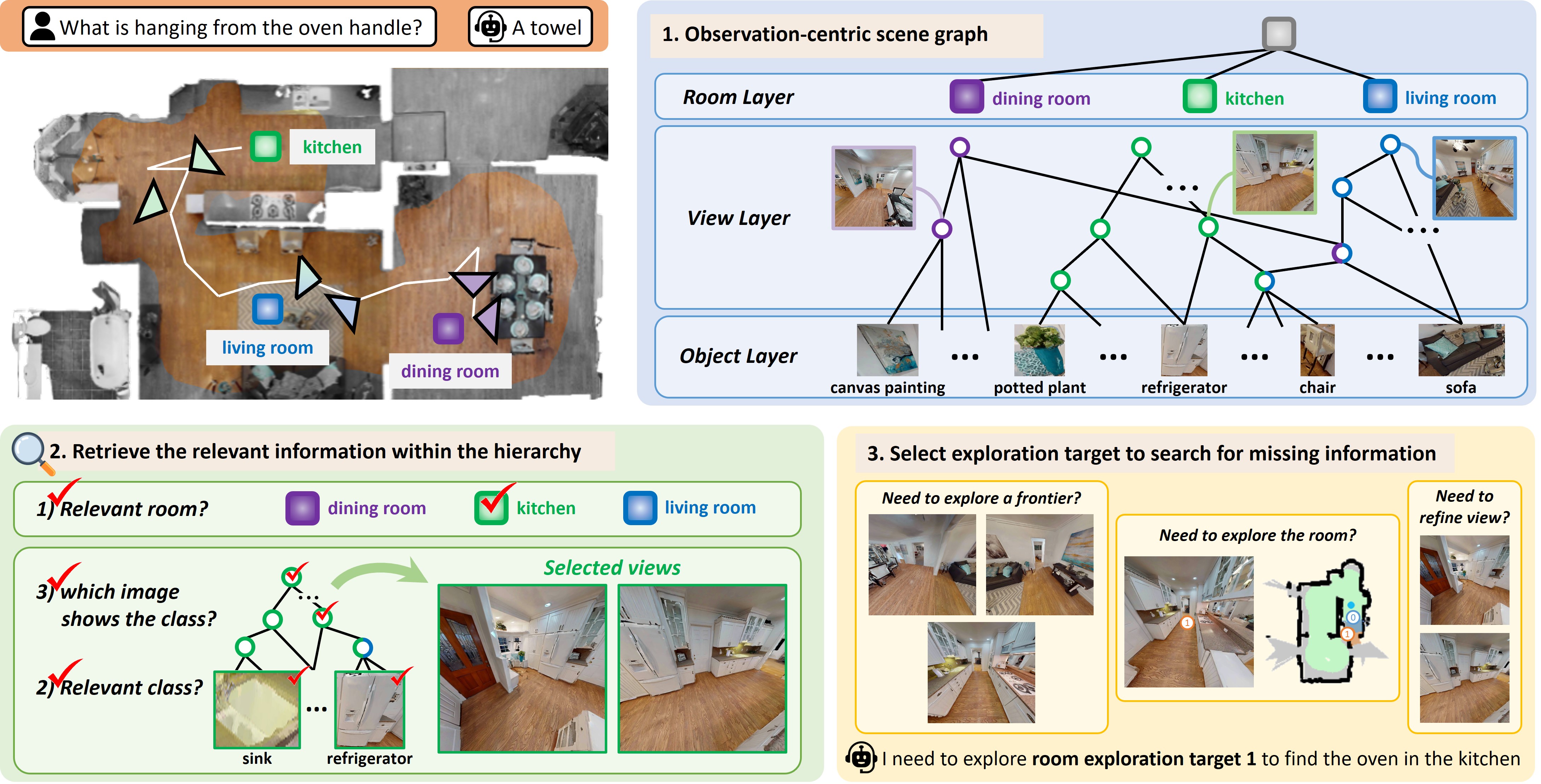}
    \caption{\label{fig:thumbnail}We propose \titleshort{}, a unified framework that tightly integrates representation, retrieval, and exploration. It maintains an observation-centric scene graph that preserves rich visual evidence while structuring it hierarchically (room–view–object) to provide both spatial context and object co-visibility cues. Given a task, relevant information is retrieved across hierarchy levels, and the retrieved context naturally induces exploration candidates at corresponding scales. In this way, exploration is naturally structured by the representation itself.}
\end{figure}

From these observations, an effective scene representation for embodied reasoning and exploration should satisfy the following properties.
(1) It must preserve sufficient evidence while remaining compact, enabling reliable reasoning.
(2) Given a task, it should allow flexible access to information at an appropriate level of abstraction and be structured to leverage contextual cues to accurately distinguish task-relevant information.
(3) It should enable reasoning about what level of additional information would be most effective for progressing toward task completion.

Dense 3D representations are a popular category and are advantageous for complex spatial understanding \cite{hu20253dllm, zhu2024llava}.
They embed visual information into their 3D geometry, preserving rich semantic information \cite{peng2023openscene, wu2021scenegraphfusion, kerr2023lerf}.
However, they lack explicit abstraction for task-conditioned access, making selective extraction of task-relevant regions nontrivial.
Consequently, reasoning often must operate over large portions of the scene, which limits scalability as exploration expands.

Scene graphs are another popular category due to their concise and structured format, which is well-suited for reasoning in complex environments \cite{saxena2024grapheqa, booker2024embodiedrag}.
Existing methods either summarize environments in an object-centric form by storing object properties and relations \cite{gu2024conceptgraphs, wald2020learning} or abstract them hierarchically \cite{hughes2022hydra, rosinol2021kimera}.
Although these representations support efficient search at appropriate abstraction levels, they may lose task-relevant evidence due to premature abstraction before the task is specified.
Recent methods instead perform reasoning directly over task-relevant image subsets \cite{yang20253d, fan2025embodied}, preserving rich visual information without early abstraction. 
While effective for retaining detailed evidence, these approaches lack an explicit structural organization.
Moreover, prior scene representations are primarily designed for storing current information about the environment, offering limited support for reasoning about what level of additional information is needed.

Motivated by the trade-off between structured scene graphs and image-based reasoning, and by the need to reason about what to observe next, we propose an observation-centric scene graph.
This representation preserves rich visual evidence and structurally grounds task-aware retrieval and exploration within a unified hierarchy.
The proposed scene graph consists of three layers: room, view, and object.
First, the room layer stores observed room types.
It organizes views by room type, providing coarse cues to disambiguate similar information in lower layers.
Second, the view layer stores selected observed images.
It provides visual context, including relationships between objects and the background where each individual object is observed.
We hypothesize that object context is determined by which other objects are observed together with it across views.
Therefore, we design a view selection method that preserves diverse object co-visibility among observed images while avoiding visual redundancy.
Third, the object layer stores object-centric crops for each object.
It complements fine-grained visual evidence that cannot be easily captured by the view layer.

Once the scene representation is updated, we perform hierarchical retrieval to assess whether sufficient task-relevant evidence (i.e., observations needed for the task) has been collected.
It identifies task-relevant evidence at the appropriate abstraction level by leveraging both spatial context in the room--object hierarchy and semantic context in the view--object hierarchy.
If evidence remains insufficient, we invoke our exploration strategy to jointly infer the missing task-relevant information and the exploration option that most effectively resolves it.
For room-level retrieval, the agent uses room-level spatial cues and associated views to infer room exploration candidates that expand task-relevant room information.
At the view level, it uses object and background context in retrieved views to infer view-refinement candidates (e.g., closer view, wider context, or different angle) that can acquire richer task-relevant evidence.
Object-level evidence refine sufficiency assessment rather than to generate exploration candidates.
Altogether, we unify representation, retrieval, and exploration by structuring the exploration option space according to the retrieved abstraction level in a single task-aware loop.
In short, our contributions are as follows.
\begin{itemize}
\item A task-agnostic, observation-centric scene graph that hierarchically organizes visual observations for task-aware reasoning and exploration
\item An informative view selection strategy for constructing a compact, context-rich view layer
\item A scene graph-guided coarse-to-fine retrieval scheme that jointly infers missing task-relevant information and effective exploration options using hierarchical structural and semantic context
\item An adaptive exploration strategy that adjusts exploration scale within the scene graph hierarchy based on identified task-relevant information gaps
\end{itemize}

\section{Related Works}

\label{sec:related_works}

\noindent\textbf{Scene Representation for Embodied Agents.}
Constructing an effective scene representation is fundamental for embodied agents, since it determines what information is preserved for reasoning and exploration.
Dense 3D maps \cite{peng2023openscene, takmaz2023openmask3d, zhang2023clip, kerr2023lerf} encode rich geometric and semantic information.
However, they are often memory-intensive and lack explicit abstraction for task-conditioned access.
As exploration expands under partial observations, these limitations hinder scalable reasoning and decision-making.
Scene graph-based representations \cite{gu2024conceptgraphs, wu2021scenegraphfusion, wald2020learning, rosinol2021kimera, hughes2022hydra} provide compact structural abstractions that improve reasoning efficiency, yet abstraction can discard visual details needed for future tasks.
This limitation is evident in OpenEQA \cite{majumdar2024openeqa}, where ConceptGraph \cite{gu2024conceptgraphs} is used as a baseline, and the abstraction can discard rich visual evidence that remains useful across tasks.
To preserve richer evidence, 3D-Mem \cite{yang20253d} maintains informative image subsets alongside structural abstractions, and \cite{wang2025expand} builds on this image-memory paradigm for task-agnostic embodied reasoning.
However, task-conditioned image memory, as in GraphEQA \cite{saxena2024grapheqa}, is less suitable for long-horizon or sequential settings where task objectives change over time.
A concurrent task-agnostic approach \cite{cai2025vision} combines scene graphs with keyframes selected by geometric coverage, but geometric coverage alone may not capture non-uniform semantic importance.
Compared with these approaches, we build a task-agnostic hierarchical memory over observations themselves and select images that preserve diverse object co-visibility while reducing redundancy.

\noindent\textbf{VLM-Based Reasoning and Exploration Policies.}
Vision-language models (VLMs) \cite{hurst2024gpt, bai2025qwen3, hu20253dllm} have become a strong foundation for embodied reasoning and exploration.
Existing VLM-based embodied methods are divided into end-to-end and modular paradigms.
End-to-end methods \cite{ramrakhya2023pirlnav, wijmans2022ver} directly predict low-level navigation actions from egocentric observations, but their generalization remains limited under current data and model scales \cite{hong2021vln, yu2023frontier}.
Modular methods decouple high-level reasoning from low-level control by using explicit intermediate representations, enabling stronger use of pretrained VLM reasoning.
Within this paradigm, some methods \cite{ren2024explore, yokoyama2024vlfm, shah2023navigation, jiang2025beyond} build task-conditioned semantic maps from observed targets to guide exploration.
Although effective for single tasks, such task-conditioned maps are inefficient for lifelong or sequential missions with changing objectives.
Other methods \cite{yang20253d, wang2025expand} maintain task-agnostic memory and delegate next-target selection to a VLM over candidate options at each step.
Importantly, the amount of information required to complete a task varies given the agent's current knowledge, making it crucial to select an appropriate exploration scope.
Nonetheless, prior work has paid limited attention to how exploration scope should be adaptively selected to resolve such insufficiency.
Fine-EQA \cite{jiang2025beyond} partially addresses this issue via room-level triggering, but it still depends on task-conditioned maps.
In contrast, we retrieve task-relevant cues from a task-agnostic hierarchical observation memory and use them to structure exploration options at different levels, enabling adaptive exploration scale control.

\section{Method}
\label{sec:method}

\begin{figure}[t]
    \centering
    \includegraphics[width=\textwidth]{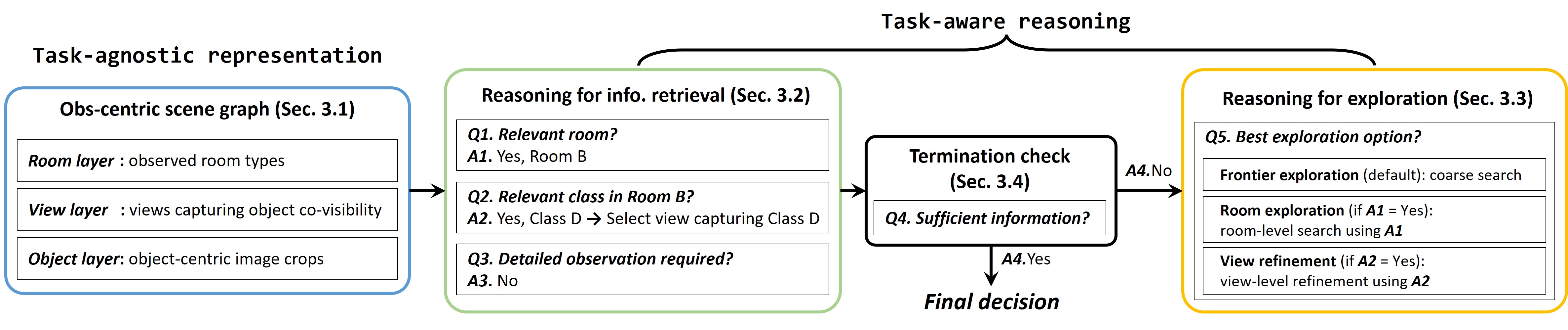}
    \caption{\label{fig:outline}The overall pipeline of the proposed unified reasoning framework, \titleshort{}.}
\end{figure}

We propose a unified reasoning framework that tightly integrates representation, retrieval, and exploration within a hierarchical observation-centric scene graph, structurally grounding task-conditioned decisions in its hierarchy. An overview of the proposed framework is illustrated in Fig.~\ref{fig:outline}. At each planning step, our system updates the proposed task-agnostic scene graph using newly acquired observations (Sec.~\ref{sec:representation}). It then performs hierarchical retrieval over the scene graph using an LLM, progressively narrowing down the search space to identify task-relevant information (Sec.~\ref{sec:retrieval}). Based on the retrieved evidence, the system determines whether the available information is sufficient to complete the task (Sec.~\ref{sec:termination}). If the information is insufficient, exploration options corresponding to the abstraction level of the retrieved task-relevant information are activated. For example, if a task-relevant room is identified, room exploration is added to the set of exploration options. The VLM then selects the most appropriate exploration target by reasoning about which option is most likely to acquire the missing information required to answer the task (Sec.~\ref{sec:exploration}).

\subsection{Observation-Centric Scene Graph}\label{sec:representation}
Our scene graph is observation-centric, meaning that we retain and hierarchically organize the visual observations themselves across multiple abstraction levels. It consists of three layers as shown in Fig.~\ref{fig:thumbnail}: a room layer that serves as a coarse spatial anchor for lower layers, a view layer that preserves object co-visibility as contextual information, and an object layer that provides fine-grained detail for each object. This representation provides a structural basis for drawing conclusions for the task or identifying which abstraction level requires additional information.

\textbf{Object layer.} Given a sequence of RGB-D images and camera poses, we first update the object set in the scene following \cite{gu2024conceptgraphs}. Each object node is defined as the object-centric crop from the image that best captures the object, preserving fine-grained visual details. From the detection results, a visibility mask $\mathcal{M}$ is updated, where $\mathcal{M}_{i,j}=1$ if the $j$-th object $o_j$ is detected in the $i$-th view $v_i$, and $0$ otherwise. Edges between the object and the view layer are defined by $\mathcal{M}$.

\textbf{View layer.} We treat object co-visibility--defined as multiple objects appearing in the same view—— as intermediate contextual information between rooms and objects. To make this layer semantically rich and compact, we propose a method that selects a subset of images that captures diverse co-visibility among objects and avoids visual redundancy. We construct a hierarchical cluster of objects using $\mathcal{M}$, and allocate an image to each node in the hierarchy. 

To promote diverse co-visibility, perform agglomerative clustering \cite{johnson1967hierarchical} based on similarity $\mathrm{Sim}(C_i,C_j)=|V(C_i\cup C_j)|/|V(C_i)\cup V(C_j)|$ where the child-aware view set $V(C)$ is defined as follows:
{\small
\begin{align}\label{eq:child_aware_view}
    V(C)=
    \begin{cases}
    \{v_i \mid \mathcal{M}_{i,j}=1 \} & \text{if } C=\{o_j\} \\
    \bigcap_k \{v_i \mid \exists o_j\in C_k : \mathcal{M}_{i,j}=1 \} & \text{otherwise}
    \end{cases}~,
\end{align}
}
where $|\cdot|$ is number of elements of set, and $C_k$ is $k$-th child of $C$. The above definition associates a view with a cluster if it observes at least one object from each child cluster. Objects can be co-visible only occasionally, and such rare co-visibility can still be informative. By re-evaluating visibility at the object level for each child cluster, we can preserve these potentially useful relationships. 

After constructing the hierarchy, we traverse it in inverse merge order and assign each cluster $C$ a representative view $v_C\in V(C)$. To balance visual diversity and information preservation, $v_C$ is selected as the view most dissimilar to previously selected views based on CLIP similarity computed over objects in $C$. If no sufficiently distinct view exists, the cluster is pruned, and its children are reattached to the parent. Inverse merge order prioritizes clusters with fewer candidate views, preserving limited contextual information before reducing redundancy among clusters with several candidate views.

During exploration, the object hierarchy and representative views are updated incrementally as new observations arrive. Newly observed objects are locally integrated into the hierarchy, preserving stable higher-level abstractions while refining lower-level clusters. Representative views are assigned for newly merged clusters following the same selection strategy, resulting in an updated view layer $\mathcal{V}$ for the current step. The incremental update process of the view layer is illustrated in Fig.~\ref{fig:viewlayer1}, and details of the incremental update are described in the supplementary material.

\begin{figure}[h]
    \centering
    \includegraphics[width=\textwidth]{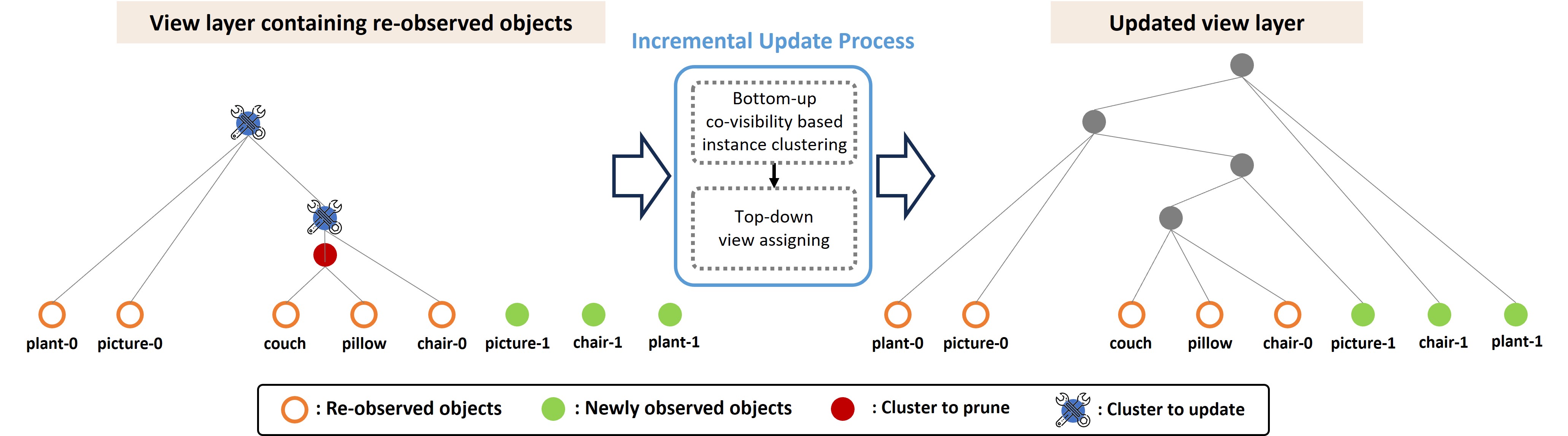}
    \caption{\label{fig:viewlayer1}Incremental update of the view layer. Given new observations, a sub-layer containing re-observed objects is first extracted from the current view layer. Newly observed and re-observed objects are then integrated by bottom-up clustering, followed by top-down view assignment, resulting in an updated view layer.}
\end{figure}

\textbf{Room layer.} The room layer represents semantic room types inferred from observations.
Every time a view $v_j$ is added to the view layer, we query a VLM to predict the types of rooms observed and the room affiliation of each detected object. To obtain robust room-object assignments $r_i$ for object $o_i$, we request VLM to output room-wise probabilities supported by the view $P(v_j|r_i=k)$ and accumulate across views via Bayesian update $P(r_i=k|v_j) = P(v_j|r_i=k) P(r_i=k) / \sum_{k'}(P(v_j|r_i=k')P(r_i=k'))$ where $k$ denotes the room index. For room-view assignments, we combine two sources: (1) direct room predictions, and (2) object-level cues, where a view is associated with a room if it contains objects assigned to that room. This conservative fusion improves robustness against noisy VLM predictions for downstream reasoning. Collectively, these layers form our observation-centric scene graph. Detailed implementation is provided in the supplementary material.

\subsection{Reasoning for Information Retrieval}\label{sec:retrieval}
While the scene graph is designed to comprehensively retain rich information about the entire environment to support a wide range of tasks, the retrieval stage filters and focuses on task-relevant information. Rather than performing a flat search over all stored observations, we design a retrieval scheme that faithfully exploits the hierarchy of our scene graph while maintaining a bounded query budget.

To retrieve information at an appropriate level of granularity while leveraging contextual cues across abstraction levels, we traverse the scene graph and progressively specify relevant information. At each layer, we query the LLM once to select up to $k$ nodes in the order of relevance to the task. The candidate space for the next layer is restricted to nodes structurally connected to the selected nodes. This design enables multi-branch reasoning without requiring repeated LLM calls for every possible path.

At the room layer, the LLM is provided with room types and the object classes assigned to each room. Depending on the task, selection may be primarily driven by explicit room-type cues or by object-level hints that imply likely room categories. At the view layer, we apply a two-stage strategy. We first ask the LLM to identify task-relevant object classes among those connected to the candidate views. We then greedily select the minimum number of views to cover these classes, until all classes are covered or $k$ views are chosen. When we queried LLM to select views in a single step, we found that the relevance reasoning was interrupted by object co-visibility information. By separating class-level semantic reasoning from view selection, our design focuses relevance estimation on object semantics while reducing redundancy among selected views. Finally, at the object layer, the LLM selects objects when focused observation of the object is assumed to be important to succeed in the task.

Through this process, we retrieve relevant room types, views, and object crops, which form a principled interface between our scene graph and downstream exploration and decision-making. We describe the prompt and details for retrieval in the supplementary.

\begin{figure}[t]
    \centering
    \includegraphics[width=\textwidth]{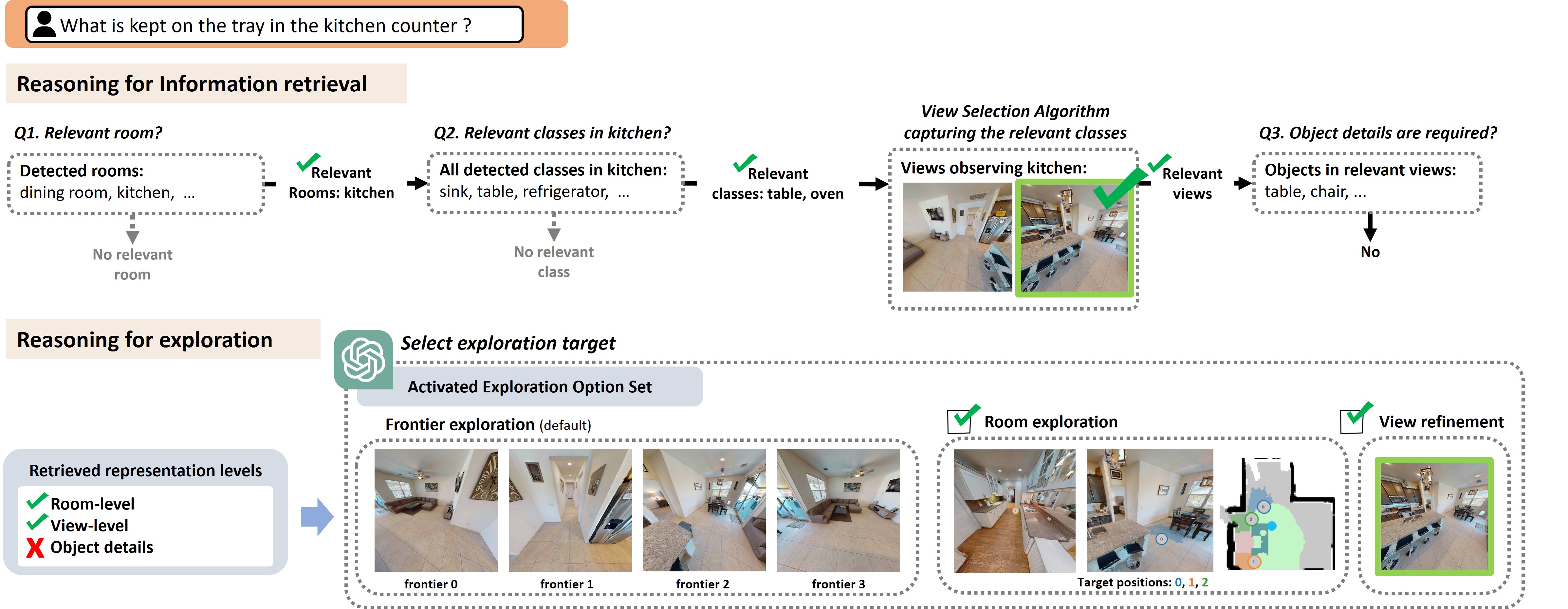}
    \caption{\label{fig:retrieval_exp}Reasoning process for information retrieval and exploration. In the information-retrieval stage, the questions denoted by \textit{Q} are queried to an LLM; through three sequential queries, the system retrieves task-relevant information and selects views that capture the relevant classes. In the exploration stage, a VLM is queried to choose an exploration option and its target. An exploration option set is activated based on the amount of retrieved evidence, and the VLM outputs the best option (frontier exploration, room exploration, or view refinement) along with the corresponding target.}
\end{figure}

\subsection{Reasoning for Exploration}\label{sec:exploration}
Our exploration strategy responds to insufficiency in the task-relevant information. Our intuition is as follows: \textit{When the available information is insufficient, each retrieved representation level serves as a cue for how it may be complemented.} For example, if a room is retrieved, unexplored regions within that room become exploration candidates. If a view is retrieved, adjusting the viewpoint around that view may complement the missing information. If no specific level provides relevant information, the agent may explore frontier regions to gather broader scene observations. Following this intuition, we design an exploration strategy in which the option space is structured according to the retrieved representation levels. Fig.~\ref{fig:retrieval_exp} illustrates how the option space is structured and presented as VLM-compatible inputs.

Importantly, because our scene graph preserves visual observations in an observation-centric hierarchy, we can construct VLM-compatible visual inputs corresponding to each exploration candidate. We instantiate three types of exploration options: room exploration, view refinement, and frontier exploration.

In addition to the scene graph, we maintain several geometric representations that support option construction. We build an occupancy map to model navigable space and support the computation of exploration target positions. Explored regions are defined as nearby areas along the agent's trajectory, while unexplored regions correspond to navigable but unvisited areas. We also maintain a room segmentation map, obtained by propagating SAM3 \cite{carion2025sam} predictions from view nodes to the top-down map.

\textbf{Room exploration.}
This option is activated when rooms are retrieved, and the agent is located in one of the retrieved rooms. We cluster unexplored regions within the room such that each cluster remains below a predefined size $r$ while minimizing the number of clusters. For each cluster, we compute a target position that enables effective observation of that region. To construct VLM-compatible inputs, we visualize exploration targets in two forms. First, we project region clusters onto views connected to the room and mark successfully projected clusters together with their corresponding target positions. We prioritize retrieved views for projection and consider additional views only when necessary. Second, we visualize all region clusters and their target positions on the occupancy map, together with explored regions and the current agent position, providing spatial context.

\textbf{View refinement.}
Retrieved views serve as refinement candidates. Although a retrieved view may observe a task-relevant region, it can still be insufficient due to (1) limited scale, (2) insufficient nearby context, or (3) suboptimal viewing angle. We define three viewpoint adjustment strategies: closer view, wider context, and different angle. Closer view increases the resolution of the task-relevant region, wider context expands the surrounding semantic context, and different angle reduces occlusion or captures an alternative geometric perspective. If the VLM selects this option, it outputs the index of the view to refine, a description of the task-relevant region being observed, and the appropriate refinement strategy. We apply SAM3 \cite{carion2025sam} to obtain a mask corresponding to the described region, and a preset function associated with the chosen strategy computes the target position to better observe the region.

\textbf{Frontier exploration.}
Frontiers serve as the default option, which are activated regardless of the retrieval result. We follow \cite{yang20253d} to update frontier regions and construct corresponding images observing them. To prevent the agent from oscillating between frontiers, we match current frontiers with previously selected ones based on their position and orientation, and structure the prompt so that a new frontier is selected only when it offers significantly greater potential to gather task-relevant information.

We restrict the VLM to output high-level exploration selection, decoupling semantic reasoning from computation of precise target coordinates, as VLMs lack accurate coordinate-level reasoning ability \cite{davoodi2025llms, mirzadeh2024gsm}. For room exploration and frontier options, target positions are precomputed during option construction. For view refinement, target positions are computed from the task-relevant region mask and the selected refinement strategy.

Overall, exploration targets are structured according to the retrieved representation hierarchy, preserving the structural grounding established in the scene graph and enabling principled, task-aware exploration. Prompt and additional details about the method are provided in the supplementary material.

\subsection{Termination Check}\label{sec:termination}
We query VLM to determine whether to stop or continue exploration based on the sufficiency of the retrieved information. We separate such reasoning from selecting the exploration option to avoid interference between the two decisions. Prompt is provided in the supplementary material.

\section{Experiments} \label{sec:exps}
\subsection{Experimental setting}\label{sec:setup}
\subsubsection{Implementation Details.}
For every benchmark, we employ GPT-4o (2024-11-20) as the VLM for memory retrieval, termination checking, exploration selection, and final answer prediction.
To construct the room layer and its hierarchy between the view and object layers, we employ GPT-4o-mini (2024-07-18).
For efficient updates of the hierarchical clustering that defines the view layer, cluster nodes with height $H\geq$ 3 (the maximum distance to descendant leaves) are fixed during updates.
During retrieval, we select up to $k$$=$$5$ relevant entries at each layer.
Once the target position is determined, we assume a collision-free trajectory planner navigates the agent to that location.
In practice, we use the built-in pathfinder in Habitat Sim, following VLM-guided exploration approaches \cite{ren2024explore, yang20253d}.
Additional implementation details are provided in the supplementary. 

\subsubsection{Benchmarks.}
We evaluate our method on three benchmarks covering different aspects of embodied reasoning and exploration.
To evaluate the ability of our scene representation and retrieval scheme, we use Episodic Memory Embodied Question Answering (EM-EQA) benchmark, a subset of OpenEQA \cite{majumdar2024openeqa}.
It contains over 1600 questions spanning recognition, functional reasoning, and spatial understanding.
Each episode provides a pre-recorded RGB-D trajectory with camera parameters, allowing evaluation without active exploration.
We further evaluate the full capability of our system on Active EQA (A-EQA), another subset of OpenEQA that additionally requires active exploration.
The benchmark consists of 557 questions derived from EM-EQA.
Following the evaluation protocol in OpenEQA and due to resource limitations, we evaluate on the official 184-question subset.
Finally, we evaluate our approach in the goal-oriented setting of GOAT-bench \cite{khanna2024goat}.
Unlike OpenEQA, GOAT-Bench focuses on navigation toward multi-modal goals, including object categories, language descriptions, and image goals.
Each episode contains multiple subtasks, emphasizing the need for task-agnostic scene representations.
Following the evaluation setup of 3D-Mem, we use a 1/10 subset of the “Val Unseen” split.

\subsubsection{Metrics.}
For both EM-EQA and A-EQA, we adopt LLM-Match proposed in OpenEQA.
For each question, GPT-4o evaluates the correctness of the predicted answer relative to the ground truth, producing a score that is normalized to a 0-100$\%$ scale.
For A-EQA, we additionally report LLM-Match SPL (LLM SPL), which measures exploration efficiency by incorporating the ratio between the predicted and reference path lengths.
For GOAT-bench, we report Success Rate (SR) and Success weighted by Path Length (SPL).
A subtask is considered successful if the agent finishes within 1 meter of the goal.

\subsubsection{Baselines.}
We primarily compare against 3D-Mem, a recent state-of-the-art approach for embodied reasoning and exploration.
We also include the original baselines provided by each benchmark.
For EM-EQA, baselines represent diverse memory designs, including abstraction-based memories (ConceptGraph Captions \cite{gu2024conceptgraphs}), dense 3D memories (Sparse Voxel Maps Captions \cite{majumdar2024openeqa}), and image-set based memories (Multi-Frame and 3D-Mem \cite{yang20253d}).
For A-EQA and GOAT-bench, baselines include both end-to-end policies (SceneAct-NN and MTU3D \cite{zhu2025move}) and VLM-guided exploration methods (Explore-EQA \cite{ren2024explore} and 3D-Mem).

\begin{table}[t]
\centering
\caption{Experiments on EM-EQA.}
\label{tab:em_result}
\resizebox{0.9\textwidth}{!}{
\begin{tabular}{lccccccc}
\toprule
 Method & Blind & CG Cap. & SVM Cap. & Frame Cap. & Multi-Frame & 3D-Mem & Ours \\
\midrule
{\bf LLM-Match} & 35.5 & 34.4 & 34.2 & 38.1 & 48.1 & 57.2 & {\bf 61.5} \\
{\bf Avg. Frames} & - & - & - & - & 3.0 & 3.1 & 3.3 \\
\bottomrule
\end{tabular}
}
\end{table}

\subsection{Main Experimental Results}\label{sec:eval_main}
\begin{figure}[t]
    \centering
    \includegraphics[width=0.8\textwidth]{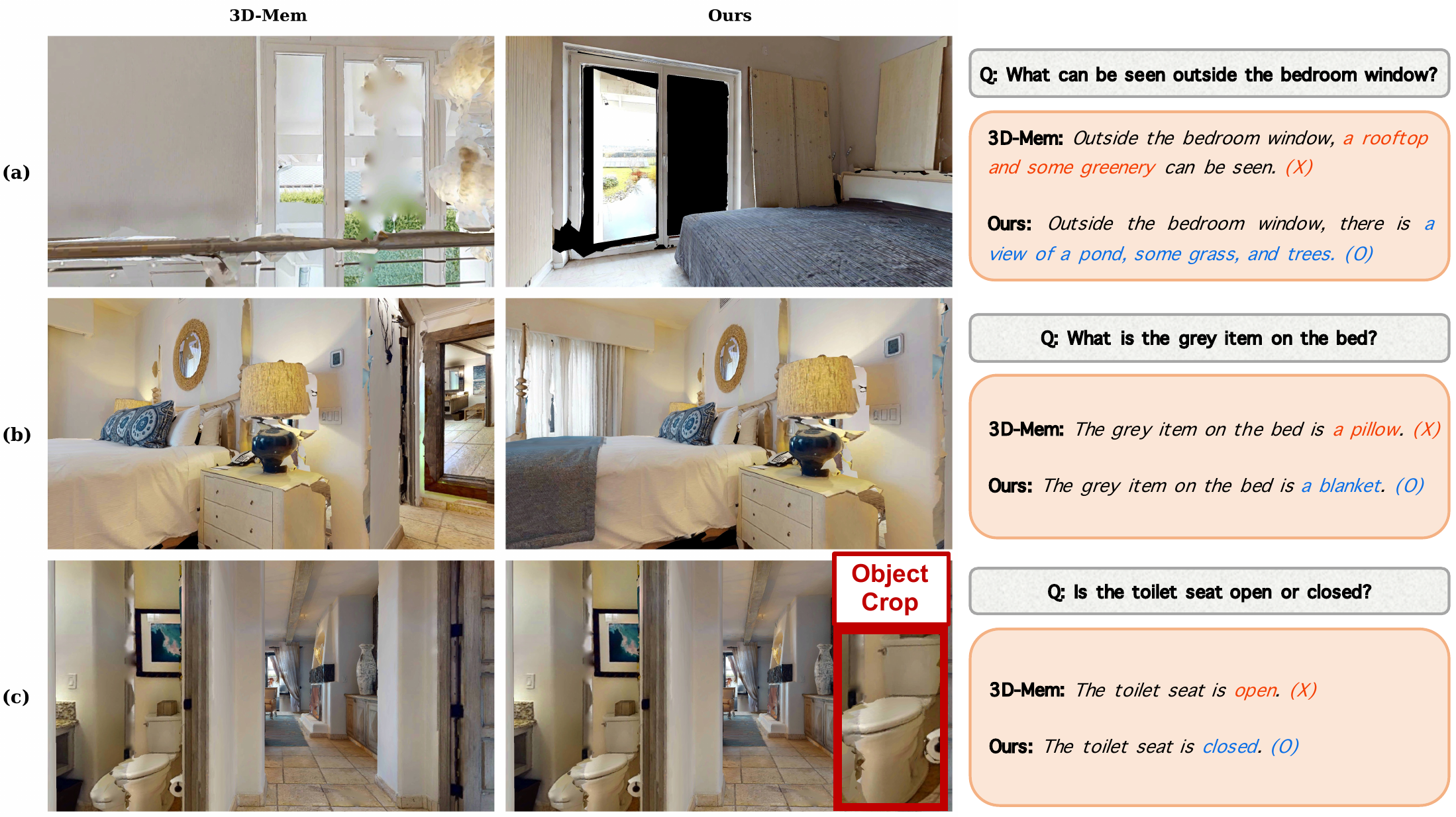}
    \caption{\label{fig:result_memory} Qualitative comparison on selected EM-EQA episodes between 3D-Mem (left)
and our method (right). Each row highlights the role of each layer in our observation-
centric hierarchy: (a) room-level retrieval narrows candidates and mitigates lexical
exact-match bias, (b) view-level memory preserves informative co-visibility patterns,
and (c) object-level crops provide fine-grained evidence for object-state reasoning.
}
\end{figure}

\subsubsection{Evaluation on Representation}


We first evaluate the effectiveness of our scene representation on EM-EQA benchmark.
As shown in Table \ref{tab:em_result}, our method outperforms all baselines, indicating that preserving both rich visual evidence and structural scene information leads to stronger performance than representations that lack either.
We also compare the number of frames queried to the VLM, which measures frame efficiency.
Our method improves by 4.3 points over 3D-Mem while increasing the average number of frames by only 0.2 per episode.

To better understand when our representation is advantageous, Fig. \ref{fig:result_memory} presents examples highlighting the role of each component.
{\bf (a) Hierarchical retrieval}. The target object (a bedroom window) is not detected, while another window appears in a different room. 3D-Mem retrieves the incorrect image due to lexical matching. In contrast, our method first retrieves the bedroom at the room level, restricting candidates and enabling the LLM to rely on contextual cues (e.g., the bed). This hierarchical filtering reduces exact-match bias.
{\bf (b) View-level memory}. 
3D-Mem selects images that cover all detected objects using the minimum number of views.
In contrast, our method retains views that capture rare co-visibility patterns, including wide observations that can be used to answer the question successfully.
{\bf (c) Object-level crops.}
The retrieved view observes a toilet from a distance, making it difficult to determine whether the seat is closed.
Our method supplements the view with an object crop from the most informative observation, enabling fine-grained reasoning about object states.

\subsubsection{Evaluation on Exploration}

We evaluate exploration performance on A-EQA by comparing two types of baselines:
(1) question-agnostic exploration methods and (2) VLM-guided exploration methods.
Since 3D-Mem relies on GPT-4o, we re-run its official code using the same GPT-4o version as our system for a fair comparison, denoted as $\dagger$.
As shown in Table~\ref{tab:aeqa_result}, our method achieves the best results in both accuracy and efficiency.
This suggests that our representation not only improves the retrieval of previously observed information but also guides exploration in a way that better aligns with VLM reasoning.

\begin{figure}[t]
    \centering
    \includegraphics[width=0.9\textwidth]{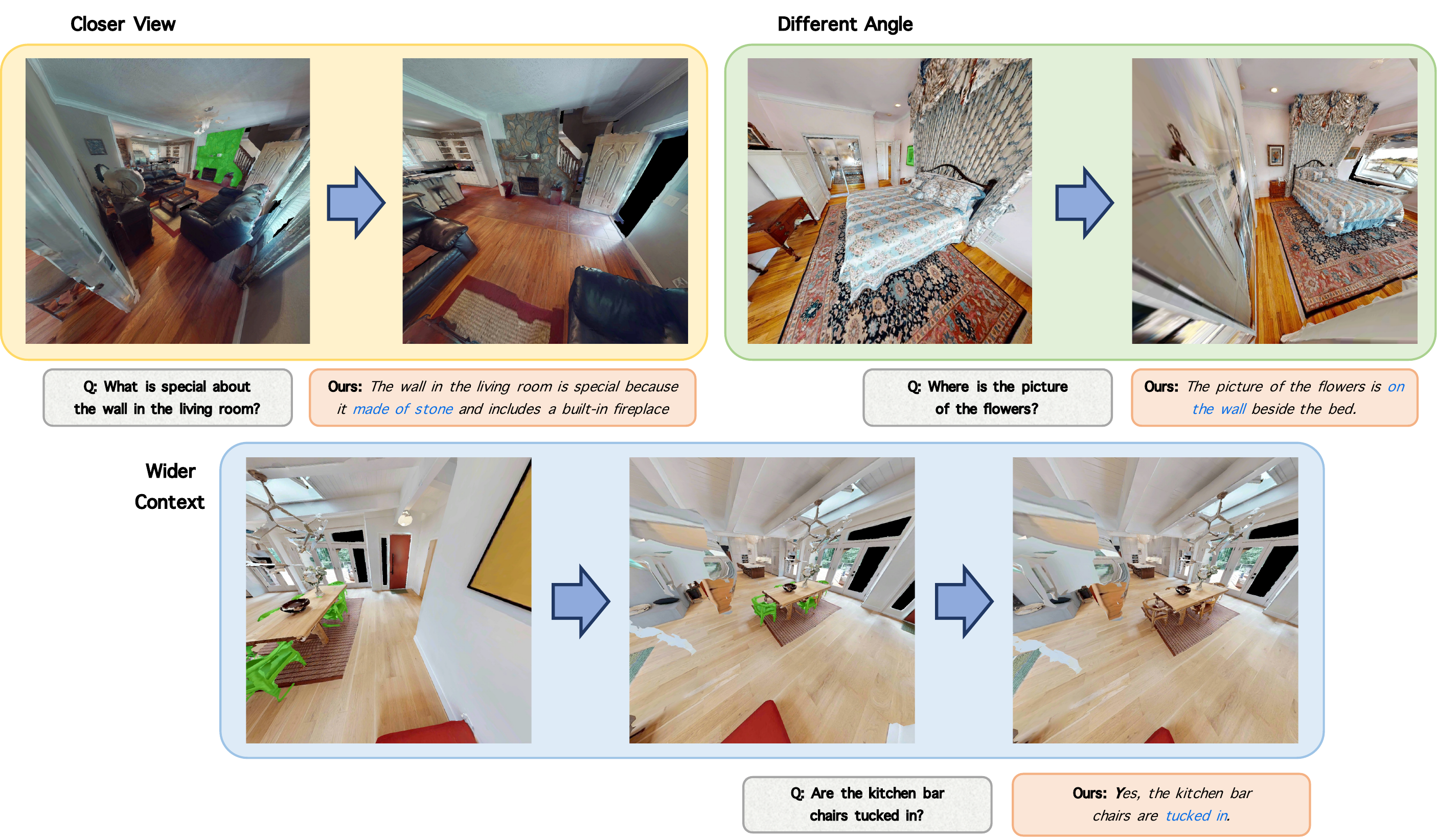}
    \caption{\label{fig:result_view_refine} Cases where each view refinement strategy correctly guides the agent.
}
\end{figure}

To further illustrate how our exploration strategies operate in practice, Fig. \ref{fig:result_view_refine} presents examples of the view refinement options. 
The regions identified by SAM are highlighted in green for visualization.
In the closer view case, the agent approaches the target. In the different angle case, the agent changes its viewing direction. In the wider context case, the agent steps back to capture the broader context.
We further analyze frame efficiency during exploration. 3D-Mem selects 10.94 images from 39.76 observations per episode and retrieves 3.26 images for VLM queries. Our method stores slightly more informative views (12.97 from 44.48 observations) but retrieves only 2.34 images on average. This demonstrates that our representation reduces the number of images actually queried to the VLM while preserving useful observations.

Finally, we evaluate on GOAT-Bench, which focuses on goal-oriented navigation.
As shown in Table \ref{tab:goat_result}, our method achieves the best performance in both SR and SPL, indicating that our approach generalizes across two embodied reasoning settings: open-ended question answering and goal-oriented navigation.

\begin{table}[t]
\centering

\begin{minipage}[t]{0.48\linewidth}
\centering
\setlength{\tabcolsep}{4pt}
\caption{Experiments on A-EQA.}
\label{tab:aeqa_result}
\resizebox{\linewidth}{!}{
\begin{tabular}{lcc}
\toprule
Method & LLM Match $\uparrow$ & LLM SPL $\uparrow$ \\
\midrule

CG Cap. \cite{majumdar2024openeqa} & 34.4 & 6.5 \\
SVM Cap. \cite{majumdar2024openeqa} & 34.2 & 6.4 \\
Frame Cap. \cite{majumdar2024openeqa} & 38.1 & 7.0 \\
Multi-Frame \cite{majumdar2024openeqa} & 41.8 & 7.5 \\

\midrule
Explore-EQA \cite{ren2024explore} & 46.9 & 23.4 \\
TANGO \cite{ziliotto2025tango} & 37.2 & -- \\
3D-Mem$\dagger$ \cite{yang20253d} & 49.6 & 37.0 \\

\midrule
Ours & {\bf 54.5} & {\bf 45.0} \\
\bottomrule
\end{tabular}
}
\end{minipage}
\hfill
\begin{minipage}[t]{0.48\linewidth}
\centering
\caption{Experiments on GOAT-Bench.}
\label{tab:goat_result}
\resizebox{0.85\linewidth}{!}{
\begin{tabular}{lcc}
\toprule
Method & SR $\uparrow$ & SPL $\uparrow$ \\
\midrule

Modular GOAT \cite{khanna2024goat} & 24.9 & 17.2 \\
Modular CoW \cite{khanna2024goat} & 16.1 & 10.4 \\
SenseAct-NN Skill Chain \cite{khanna2024goat} & 29.5 & 11.3 \\
SenseAct-NN Monolithic \cite{khanna2024goat} & 12.3 & 6.8 \\

\midrule
Explore-EQA \cite{ren2024explore} & 55.0 & 37.9 \\
TANGO \cite{ziliotto2025tango} & 32.1 & 16.5 \\
MTU3D \cite{zhu2025move} & 47.7 & 27.7 \\
3D-Mem \cite{yang20253d} & 69.1 & 48.9 \\

\midrule
Ours & \textbf{72.7} & \textbf{51.5} \\
\bottomrule
\end{tabular}
}
\end{minipage}

\end{table}

\subsection{Ablation Studies}\label{sec:ablation}
\subsubsection{Ablation on Representation.} To quantify the contribution of each layer, we conduct ablation experiments on EM-EQA, where all methods observe the same trajectories. 
We consider three settings: (R1) Remove the hierarchy and retain only the view layer. (R2) Replace our view construction with the informative-view selection strategy of 3D-Mem \cite{yang20253d}. (R3) Remove the instance layer.
Evaluation is conducted on a 300-episode subset sampled uniformly across question categories.
Results are shown in Table \ref{tab:em_ablation_repr}.
Replacing our view construction (R2) results in the largest performance drop despite using the fewest images, highlighting the importance of view selection.
Overall, the gap between our full model and each ablation confirms that each layer contributes to the performance.

\subsubsection{Ablation on Exploration.} We also analyze the effect of each exploration option on the A-EQA 184 subset. Two variants are considered: (E1) removing the room exploration and (E2) the view refinement option. As shown in Table.~\ref{tab:a_ablation_explore}, each option contributes to both accuracy and efficiency. View refinement allows the agent to move to new viewpoints that better observe a candidate region and verify whether it matches the target. Without this option, the agent may fail to confirm potential targets and continue exploring unnecessarily. Room exploration acts as an intermediate exploration scale between frontier exploration and view refinement. It enables the agent to gather more information within a region before committing to fine-grained verification or broader exploration, improving both accuracy and efficiency.

\begin{table}[t]
\centering

\begin{minipage}[t]{0.48\linewidth}
\centering
\caption{Ablation on Representation.}
\label{tab:em_ablation_repr}
\setlength{\tabcolsep}{4pt}
\resizebox{\linewidth}{!}{
\begin{tabular}{lcccc}
\toprule
Method & (R1) & (R2) & (R3) & Ours \\
\midrule
{ LLM Match} $\uparrow$ & 58.1 & 54.2 & 59.0 & {\bf 63.5} \\
{ Avg. Frames} & 3.17 & 2.64 & 3.01 & 3.22 \\
\bottomrule
\end{tabular}
}
\end{minipage}
\hfill
\begin{minipage}[t]{0.48\linewidth}
\centering
\caption{Ablation on Exploration.}
\label{tab:a_ablation_explore}
\setlength{\tabcolsep}{4pt}

\resizebox{0.85\linewidth}{!}{ 
\begin{tabular}{lccc}
\toprule
Method & (E1) & (E2) & Ours \\
\midrule
{ LLM Match} $\uparrow$ & 52.2 & 52.7 & {\bf 54.5} \\
{ LLM SPL} $\uparrow$ & 43.7 & 42.5 & {\bf 45.0} \\
\bottomrule
\end{tabular}
}
\end{minipage}

\end{table}

\vspace{-3mm}
\section{Conclusion and Limitations} \label{sec:conclusion}
\vspace{-2mm}
We presented \titleshort{}, a unified framework that tightly integrates scene representation, information retrieval, and exploration for embodied reasoning tasks.
Our approach maintains an observation-centric hierarchy that organizes visual information into rooms, views, and object layers, preserving rich visual and structural information of the scene.
On top of this representation, we perform hierarchical retrieval to access task-relevant information.
The retrieval naturally structures the exploration candidate space, which consists of room exploration, view refinement, and frontier exploration, enabling the agent to adaptively select an exploration option.
Through this design, representation, retrieval, and exploration operate within a single task-aware loop.
Extensive experiments across multiple embodied reasoning and exploration benchmarks show that our method can accurately interpret observations and effectively explore to gather task-relevant information.
Nevertheless, our approach relies on VLM reasoning to interpret retrieved observations and select exploration options, which may be sensitive to prediction errors in perception or language understanding. 
Future work could investigate more robust integration of geometric reasoning and extend the framework to more dynamic environments and longer-horizon tasks.



\clearpage
\bibliographystyle{splncs04}
\bibliography{main}

@article{johnson1967hierarchical,
  title={Hierarchical clustering schemes},
  author={Johnson, Stephen C},
  journal={Psychometrika},
  volume={32},
  number={3},
  pages={241--254},
  year={1967},
  publisher={Springer-Verlag}
}

@inproceedings{gu2024conceptgraphs,
  title={Conceptgraphs: Open-vocabulary 3d scene graphs for perception and planning},
  author={Gu, Qiao and Kuwajerwala, Ali and Morin, Sacha and Jatavallabhula, Krishna Murthy and Sen, Bipasha and Agarwal, Aditya and Rivera, Corban and Paul, William and Ellis, Kirsty and Chellappa, Rama and others},
  booktitle={2024 IEEE International Conference on Robotics and Automation (ICRA)},
  pages={5021--5028},
  year={2024},
  organization={IEEE}
}

@article{carion2025sam,
  title={Sam 3: Segment anything with concepts},
  author={Carion, Nicolas and Gustafson, Laura and Hu, Yuan-Ting and Debnath, Shoubhik and Hu, Ronghang and Suris, Didac and Ryali, Chaitanya and Alwala, Kalyan Vasudev and Khedr, Haitham and Huang, Andrew and others},
  journal={arXiv preprint arXiv:2511.16719},
  year={2025}
}

@inproceedings{yang20253d,
  title={3D-mem: 3D scene memory for embodied exploration and reasoning},
  author={Yang, Yuncong and Yang, Han and Zhou, Jiachen and Chen, Peihao and Zhang, Hongxin and Du, Yilun and Gan, Chuang},
  booktitle={Proceedings of the Computer Vision and Pattern Recognition Conference},
  pages={17294--17303},
  year={2025}
}

@inproceedings{davoodi2025llms,
  title={Llms are not intelligent thinkers: Introducing mathematical topic tree benchmark for comprehensive evaluation of llms},
  author={Davoodi, Arash Gholami and Davoudi, Seyed Pouyan Mousavi and Pezeshkpour, Pouya},
  booktitle={Proceedings of the 2025 Conference of the Nations of the Americas Chapter of the Association for Computational Linguistics: Human Language Technologies (Volume 1: Long Papers)},
  pages={3127--3140},
  year={2025}
}

@article{mirzadeh2024gsm,
  title={Gsm-symbolic: Understanding the limitations of mathematical reasoning in large language models},
  author={Mirzadeh, Iman and Alizadeh, Keivan and Shahrokhi, Hooman and Tuzel, Oncel and Bengio, Samy and Farajtabar, Mehrdad},
  journal={arXiv preprint arXiv:2410.05229},
  year={2024}
}

@article{rosinol2021kimera,
  title={Kimera: From SLAM to spatial perception with 3D dynamic scene graphs},
  author={Rosinol, Antoni and Violette, Andrew and Abate, Marcus and Hughes, Nathan and Chang, Yun and Shi, Jingnan and Gupta, Arjun and Carlone, Luca},
  journal={The International Journal of Robotics Research},
  volume={40},
  number={12-14},
  pages={1510--1546},
  year={2021},
  publisher={SAGE Publications Sage UK: London, England}
}

@article{hughes2022hydra,
  title={Hydra: A real-time spatial perception system for 3D scene graph construction and optimization},
  author={Hughes, Nathan and Chang, Yun and Carlone, Luca},
  journal={arXiv preprint arXiv:2201.13360},
  year={2022}
}

@inproceedings{peng2023openscene,
  title={Openscene: 3d scene understanding with open vocabularies},
  author={Peng, Songyou and Genova, Kyle and Jiang, Chiyu and Tagliasacchi, Andrea and Pollefeys, Marc and Funkhouser, Thomas and others},
  booktitle={Proceedings of the IEEE/CVF conference on computer vision and pattern recognition},
  pages={815--824},
  year={2023}
}

@article{takmaz2023openmask3d,
  title={Openmask3d: Open-vocabulary 3d instance segmentation},
  author={Takmaz, Ay{\c{c}}a and Fedele, Elisabetta and Sumner, Robert W and Pollefeys, Marc and Tombari, Federico and Engelmann, Francis},
  journal={arXiv preprint arXiv:2306.13631},
  year={2023}
}

@inproceedings{zhang2023clip,
  title={Clip-fo3d: Learning free open-world 3d scene representations from 2d dense clip},
  author={Zhang, Junbo and Dong, Runpei and Ma, Kaisheng},
  booktitle={Proceedings of the IEEE/CVF international conference on computer vision},
  pages={2048--2059},
  year={2023}
}

@inproceedings{wald2020learning,
  title={Learning 3d semantic scene graphs from 3d indoor reconstructions},
  author={Wald, Johanna and Dhamo, Helisa and Navab, Nassir and Tombari, Federico},
  booktitle={Proceedings of the IEEE/CVF Conference on Computer Vision and Pattern Recognition},
  pages={3961--3970},
  year={2020}
}

@inproceedings{wu2021scenegraphfusion,
  title={Scenegraphfusion: Incremental 3d scene graph prediction from rgb-d sequences},
  author={Wu, Shun-Cheng and Wald, Johanna and Tateno, Keisuke and Navab, Nassir and Tombari, Federico},
  booktitle={Proceedings of the IEEE/CVF Conference on Computer Vision and Pattern Recognition},
  pages={7515--7525},
  year={2021}
}

@inproceedings{khanna2024goat,
  title={Goat-bench: A benchmark for multi-modal lifelong navigation},
  author={Khanna, Mukul and Ramrakhya, Ram and Chhablani, Gunjan and Yenamandra, Sriram and Gervet, Theophile and Chang, Matthew and Kira, Zsolt and Chaplot, Devendra Singh and Batra, Dhruv and Mottaghi, Roozbeh},
  booktitle={Proceedings of the IEEE/CVF Conference on Computer Vision and Pattern Recognition},
  pages={16373--16383},
  year={2024}
}

@inproceedings{yokoyama2024vlfm,
  title={Vlfm: Vision-language frontier maps for zero-shot semantic navigation},
  author={Yokoyama, Naoki and Ha, Sehoon and Batra, Dhruv and Wang, Jiuguang and Bucher, Bernadette},
  booktitle={2024 IEEE International Conference on Robotics and Automation (ICRA)},
  pages={42--48},
  year={2024},
  organization={IEEE}
}

@article{wang2025expand,
  title={Expand Your SCOPE: Semantic Cognition over Potential-Based Exploration for Embodied Visual Navigation},
  author={Wang, Ningnan and Chen, Weihuang and Chen, Liming and Ji, Haoxuan and Guo, Zhongyu and Zhang, Xuchong and Sun, Hongbin},
  journal={arXiv preprint arXiv:2511.08935},
  year={2025}
}

@inproceedings{majumdar2024openeqa,
  title={Openeqa: Embodied question answering in the era of foundation models},
  author={Majumdar, Arjun and Ajay, Anurag and Zhang, Xiaohan and Putta, Pranav and Yenamandra, Sriram and Henaff, Mikael and Silwal, Sneha and Mcvay, Paul and Maksymets, Oleksandr and Arnaud, Sergio and others},
  booktitle={Proceedings of the IEEE/CVF conference on computer vision and pattern recognition},
  pages={16488--16498},
  year={2024}
}

@article{ren2024explore,
  title={Explore until confident: Efficient exploration for embodied question answering},
  author={Ren, Allen Z and Clark, Jaden and Dixit, Anushri and Itkina, Masha and Majumdar, Anirudha and Sadigh, Dorsa},
  journal={arXiv preprint arXiv:2403.15941},
  year={2024}
}

@article{saxena2024grapheqa,
  title={Grapheqa: Using 3d semantic scene graphs for real-time embodied question answering},
  author={Saxena, Saumya and Buchanan, Blake and Paxton, Chris and Liu, Peiqi and Chen, Bingqing and Vaskevicius, Narunas and Palmieri, Luigi and Francis, Jonathan and Kroemer, Oliver},
  journal={arXiv preprint arXiv:2412.14480},
  year={2024}
}

@article{cai2025vision,
  title={Vision to Geometry: 3D Spatial Memory for Sequential Embodied MLLM Reasoning and Exploration},
  author={Cai, Zhongyi and Du, Yi and Wang, Chen and Kong, Yu},
  journal={arXiv preprint arXiv:2512.02458},
  year={2025}
}

@inproceedings{jiang2025beyond,
  title={Beyond the destination: A novel benchmark for exploration-aware embodied question answering},
  author={Jiang, Kaixuan and Liu, Yang and Chen, Weixing and Luo, Jingzhou and Chen, Ziliang and Pan, Ling and Li, Guanbin and Lin, Liang},
  booktitle={Proceedings of the IEEE/CVF International Conference on Computer Vision},
  pages={9091--9101},
  year={2025}
}

@article{hu20253dllm,
  title={3dllm-mem: Long-term spatial-temporal memory for embodied 3d large language model},
  author={Hu, Wenbo and Hong, Yining and Wang, Yanjun and Gao, Leison and Wei, Zibu and Yao, Xingcheng and Peng, Nanyun and Bitton, Yonatan and Szpektor, Idan and Chang, Kai-Wei},
  journal={arXiv preprint arXiv:2505.22657},
  year={2025}
}

@inproceedings{fan2025embodied,
  title={Embodied videoagent: Persistent memory from egocentric videos and embodied sensors enables dynamic scene understanding},
  author={Fan, Yue and Ma, Xiaojian and Su, Rongpeng and Guo, Jun and Wu, Rujie and Chen, Xi and Li, Qing},
  booktitle={Proceedings of the IEEE/CVF International Conference on Computer Vision},
  pages={6342--6352},
  year={2025}
}

@inproceedings{zhu2025move,
  title={Move to understand a 3d scene: Bridging visual grounding and exploration for efficient and versatile embodied navigation},
  author={Zhu, Ziyu and Wang, Xilin and Li, Yixuan and Zhang, Zhuofan and Ma, Xiaojian and Chen, Yixin and Jia, Baoxiong and Liang, Wei and Yu, Qian and Deng, Zhidong and others},
  booktitle={Proceedings of the IEEE/CVF International Conference on Computer Vision},
  pages={8120--8132},
  year={2025}
}

@inproceedings{ramrakhya2023pirlnav,
  title={Pirlnav: Pretraining with imitation and rl finetuning for objectnav},
  author={Ramrakhya, Ram and Batra, Dhruv and Wijmans, Erik and Das, Abhishek},
  booktitle={Proceedings of the IEEE/CVF Conference on Computer Vision and Pattern Recognition},
  pages={17896--17906},
  year={2023}
}

@article{wijmans2022ver,
  title={Ver: Scaling on-policy rl leads to the emergence of navigation in embodied rearrangement},
  author={Wijmans, Erik and Essa, Irfan and Batra, Dhruv},
  journal={Advances in Neural Information Processing Systems},
  volume={35},
  pages={7727--7740},
  year={2022}
}

@inproceedings{hong2021vln,
  title={Vln bert: A recurrent vision-and-language bert for navigation},
  author={Hong, Yicong and Wu, Qi and Qi, Yuankai and Rodriguez-Opazo, Cristian and Gould, Stephen},
  booktitle={Proceedings of the IEEE/CVF conference on Computer Vision and Pattern Recognition},
  pages={1643--1653},
  year={2021}
}

@article{yu2023frontier,
  title={Frontier semantic exploration for visual target navigation},
  author={Yu, Bangguo and Kasaei, Hamidreza and Cao, Ming},
  journal={arXiv preprint arXiv:2304.05506},
  year={2023}
}

@inproceedings{shah2023navigation,
  title={Navigation with large language models: Semantic guesswork as a heuristic for planning},
  author={Shah, Dhruv and Equi, Michael Robert and Osi{\'n}ski, B{\l}a{\.z}ej and Xia, Fei and Ichter, Brian and Levine, Sergey},
  booktitle={Conference on Robot Learning},
  pages={2683--2699},
  year={2023},
  organization={PMLR}
}

@article{booker2024embodiedrag,
  title={Embodiedrag: Dynamic 3d scene graph retrieval for efficient and scalable robot task planning},
  author={Booker, Meghan and Byrd, Grayson and Kemp, Bethany and Schmidt, Aurora and Rivera, Corban},
  journal={arXiv preprint arXiv:2410.23968},
  year={2024}
}

@article{zhu2024llava,
  title={Llava-3d: A simple yet effective pathway to empowering lmms with 3d-awareness},
  author={Zhu, Chenming and Wang, Tai and Zhang, Wenwei and Pang, Jiangmiao and Liu, Xihui},
  journal={arXiv preprint arXiv:2409.18125},
  year={2024}
}

@inproceedings{li2022blip,
  title={Blip: Bootstrapping language-image pre-training for unified vision-language understanding and generation},
  author={Li, Junnan and Li, Dongxu and Xiong, Caiming and Hoi, Steven},
  booktitle={International conference on machine learning},
  pages={12888--12900},
  year={2022},
  organization={PMLR}
}

@inproceedings{antol2015vqa,
  title={Vqa: Visual question answering},
  author={Antol, Stanislaw and Agrawal, Aishwarya and Lu, Jiasen and Mitchell, Margaret and Batra, Dhruv and Zitnick, C Lawrence and Parikh, Devi},
  booktitle={Proceedings of the IEEE international conference on computer vision},
  pages={2425--2433},
  year={2015}
}

@article{ahn2022can,
  title={Do as i can, not as i say: Grounding language in robotic affordances},
  author={Ahn, Michael and Brohan, Anthony and Brown, Noah and Chebotar, Yevgen and Cortes, Omar and David, Byron and Finn, Chelsea and Fu, Chuyuan and Gopalakrishnan, Keerthana and Hausman, Karol and others},
  journal={arXiv preprint arXiv:2204.01691},
  year={2022}
}

@article{brohan2022rt,
  title={Rt-1: Robotics transformer for real-world control at scale},
  author={Brohan, Anthony and Brown, Noah and Carbajal, Justice and Chebotar, Yevgen and Dabis, Joseph and Finn, Chelsea and Gopalakrishnan, Keerthana and Hausman, Karol and Herzog, Alex and Hsu, Jasmine and others},
  journal={arXiv preprint arXiv:2212.06817},
  year={2022}
}

@inproceedings{zitkovich2023rt,
  title={Rt-2: Vision-language-action models transfer web knowledge to robotic control},
  author={Zitkovich, Brianna and Yu, Tianhe and Xu, Sichun and Xu, Peng and Xiao, Ted and Xia, Fei and Wu, Jialin and Wohlhart, Paul and Welker, Stefan and Wahid, Ayzaan and others},
  booktitle={Conference on Robot Learning},
  pages={2165--2183},
  year={2023},
  organization={PMLR}
}

@inproceedings{kerr2023lerf,
  title={Lerf: Language embedded radiance fields},
  author={Kerr, Justin and Kim, Chung Min and Goldberg, Ken and Kanazawa, Angjoo and Tancik, Matthew},
  booktitle={Proceedings of the IEEE/CVF international conference on computer vision},
  pages={19729--19739},
  year={2023}
}

@article{hurst2024gpt,
  title={Gpt-4o system card},
  author={Hurst, Aaron and Lerer, Adam and Goucher, Adam P and Perelman, Adam and Ramesh, Aditya and Clark, Aidan and Ostrow, AJ and Welihinda, Akila and Hayes, Alan and Radford, Alec and others},
  journal={arXiv preprint arXiv:2410.21276},
  year={2024}
}

@article{bai2025qwen3,
  title={Qwen3-vl technical report},
  author={Bai, Shuai and Cai, Yuxuan and Chen, Ruizhe and Chen, Keqin and Chen, Xionghui and Cheng, Zesen and Deng, Lianghao and Ding, Wei and Gao, Chang and Ge, Chunjiang and others},
  journal={arXiv preprint arXiv:2511.21631},
  year={2025}
}

@inproceedings{ziliotto2025tango,
  title={Tango: training-free embodied ai agents for open-world tasks},
  author={Ziliotto, Filippo and Campari, Tommaso and Serafini, Luciano and Ballan, Lamberto},
  booktitle={Proceedings of the Computer Vision and Pattern Recognition Conference},
  pages={24603--24613},
  year={2025}
}

%
%

\end{document}